\documentclass[sigconf]{acmart}

\setcopyright{none}
\renewcommand\footnotetextcopyrightpermission[1]{}
\AtBeginDocument{%
  \providecommand\BibTeX{{%
    \normalfont B\kern-0.5em{\scshape i\kern-0.25em b}\kern-0.8em\TeX}}}

\copyrightyear{2020}
\acmYear{2020}
\setcopyright{acmcopyright}\acmConference[GeoHumanities'20]{4th ACM SIGSPATIAL Workshop on Geospatial Humanities }{November 3--6, 2020}{Seattle, WA, USA}
\acmBooktitle{4th ACM SIGSPATIAL Workshop on Geospatial Humanities (GeoHumanities'20), November 3--6, 2020, Seattle, WA, USA}
\acmPrice{15.00}
\acmDOI{10.1145/3423337.3429435}
\acmISBN{978-1-4503-8163-5/20/11}

\begin{document}

\title{Normalization of Different Swedish Dialects Spoken in Finland}

\author{Mika Hämäläinen}
\email{mika@rootroo.com}
\affiliation{%
  \institution{University of Helsinki and Rootroo}
  \city{Helsinki}
  \country{Finland}
}
\author{Niko Partanen}
\email{niko.partanen@helsinki.fi}
\affiliation{%
  \institution{University of Helsinki}
  \city{Helsinki}
  \country{Finland}
}
\author{Khalid Alnajjar}
\email{khalid@rootroo.com}
\affiliation{%
  \institution{University of Helsinki and Rootroo}
  \city{Helsinki}
  \country{Finland}
}

\renewcommand{\shortauthors}{Hämäläinen, Partanen \& Alnajjar}

\begin{abstract}
  Our study presents a dialect normalization method for different Finland Swedish dialects covering six regions. We tested 5 different models, and the best model improved the word error rate from 76.45 to 28.58. Contrary to results reported in earlier research on Finnish dialects, we found that training the model with one word at a time gave best results. We believe this is due to the size of the training data available for the model. Our models are accessible as a Python package. The study provides important information about the adaptability of these methods in different contexts, and gives important baselines for further study. 
\end{abstract}




\maketitle

\section{Introduction}

Swedish is a minority language in Finland and it is the second official language of the country. 
Finland Swedish is very different from Swedish spoken in Sweden in terms of pronunciation, vocabulary and some parts of the grammar. 
The Swedish dialects in Finland also differ radically from one another and from one region to another. 
Because of the wide geographical span of the Swedish speaking communities in Finland and low population density of the country, the dialects have not remained similar.  Despite its official status, Finland Swedish has hardly received any research attention within the natural language processing community. 

This paper introduces a method for dialect transcript normalization, which enables the possibility to use existing NLP tools targeted for normative Swedish on these materials. Previous work conducted in English data indicates that normalization is a viable way of improving the accuracy of NLP methods such as POS tagging \cite{van2017normalize}. This is an important motivation as the non-standard colloquial Swedish is the language of communication on a multitude of internet platforms ranging from social media to forums and blogs. In its linguistic form, the colloquial dialectal Finland Swedish deviates greatly from the standard normative Swedish and the dialectal variants of the language spoken in Sweden, a fact that lowers the performance of the existing NLP tools for processing Swedish on such text. 
Finland Swedish is also a continuous target of research by the non-computational linguistics community, among other fields of research, and we better methods to analyze these texts are also beneficial in these academic domains.  

\begin{figure}[h]
  \centering
  \includegraphics[width=\linewidth]{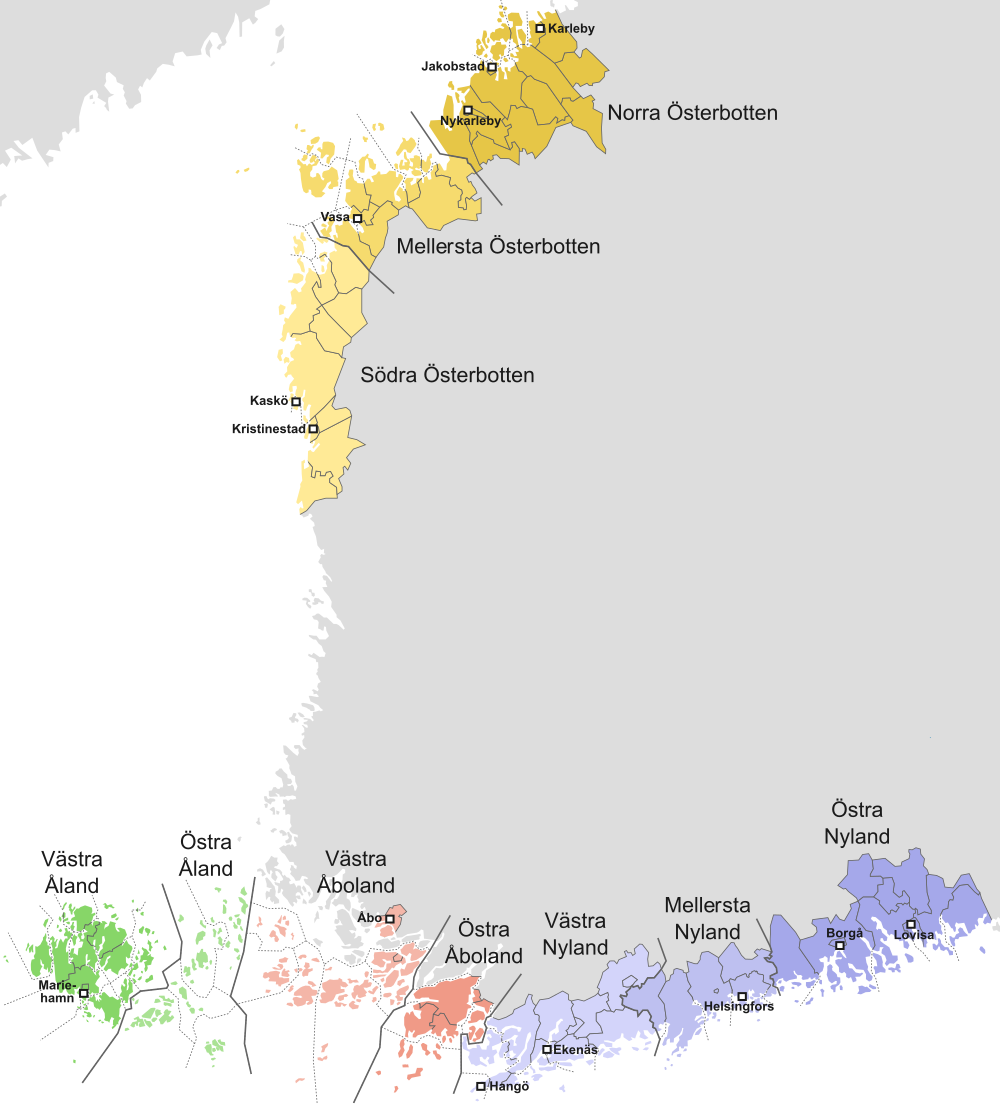}
  \caption{Dialects of Swedish in Finland. \&
    Fenn-O-maniC [CC-BY-SA], via Wikimedia
    Commons. (\url{https://commons.wikimedia.org/wiki/File:Svenska_dialekter_i_Finland.svg}).}
  \Description{Dialects of Swedish in Finland.}
\label{map}
\end{figure}

We train several dialect normalization models to cover dialects of six different Swedish speaking regions of Finland. In terms of area, this covers the Aland island and the regions along the Baltic Sea coastline that have the largest number of Swedish speaking population. See the Figure~\ref{map} for a map that shows the geographical extent. The dialect normalization models have been made available for everyone through a Python library called Murre\footnote{https://github.com/mikahama/murre}. This is important so that people both inside and outside of academia can use the normalization models easily on their data processing pipelines.

\section{Related Work}

Text normalization has been studied often in the past. 
The main areas of application have been historical text normalization, dialect text normalization, and noisy user generated text normalization. 
All these are important domains, with their own distinct challenges. 
Most important for our task is dialectal text normalization, but for the sake of thoroughness we discuss the related work in a somewhat wider context. 

Bollmann \cite{bollmann-2019-large} has provided a meta-analysis where contemporary approaches are divided into five categories: substitution lists like VARD \cite{rayson2005vard} and Norma \cite{bollmann2012-semi}, rule-based methods \cite{baron2008vard2,porta2013edit}, edit distance based approaches \cite{hauser2007unsupervised,amoia2013using}, statistical methods and most recently neural methods.

Statistical machine translation (SMT) approaches have generally been at the core of the most commonly used statistical methods. In these methods, the normalization is often assimilated with the regular translation process by training an SMT model on a character level. 
These methods have been used both for historical text \cite{pettersson2013smt, hamalainen2018normalizing, hamalainen2019paft} and contemporary dialect normalization \cite{swissgerman}.




Currently many normalization methods have utilized neural machine translation (NMT), in a way that is comparable to the earlier SMT based approaches.  \citet{bollmann-sogaard-2016-improving} used a bidirectional long short-term memory (bi-LSTM) deep neural network in a character level normalization of historical German. 
They also tested the efficiency of multi-task learning, when additional data is used during the training phase. 
In these experiments neural network approaches gave better results than conditional random fields (CRF) and Norma, whereas multi-task learning provided the best accuracy.  

An approach based on LSTMs and the noisy channel model (NCM) was tested by \citet{tursun-cakici-2017-noisy} to normalize Uyghur text. 
They used a small base dataset of $\approx$~200 sentences, which were obtained from social networks and normalized. 
The authors generated noisy synthetic data by inserting random errors into online-crawled data. 
These methods were able to normalize the text with high accuracy. 
\citet{mandal-nanmaran-2018-normalization} used an LSTM network to normalize code-mixed data, and achieved an accuracy of 90.27\%, which can be considered successful.

Within the historical text normalization, a recent study \cite{hamalainen-etal-2019-revisiting} compared various LSTM architectures, and found that bi-directional recurrent neural networks (BRNN) were more accurate than one-directional RNNs. 
Different attention models or deeper architectures did not improve the results further. 
In the same vein, additional metadata decreased the accuracy. 
For both historical and dialectal data such metadata is often available, which makes these experiments particularly relevant. 
These findings suggest that at the moment post-processing appears as the best way to improve a character level NMT normalization model. 
In this context some attention needs to be paid over the fact that dialect normalization is in many ways a distinct task from historical text normalization, although it shares similarities. 

Closely related to the current work, a very effective method has been proposed for normalization of dialectal Finnish \cite{partanen2019dialect}. 
The authors trained a character-based normalization model based on bi-directional recurrent neural network.  
In this study the training corpus was very large, almost 700,000 normalized word tokens. 

\section{Data and Preprocessing}

As our dataset, we use the collection of recordings of Finland Swedish collected between 2005 and 2008 \cite{ruotsinmurteet}. We crawl the version of this dataset that is hosted online on Finna by Society of Swedish Literature in Finland, and is CC-BY licensed\footnote{https://sls.finna.fi/Collection/sls.SLS+2098/}. This dataset consists of interviews with Swedish speaking people from different parts of Finland together with coordinates indicating where the data was collected from. The recordings have been transcribed by hand to a textual format that follows the pronunciation of the participants. These transcriptions have also been normalized by hand to standard Finland Swedish writing. These recordings represent dialects of six regions of Finland: \textit{Åland}, \textit{Åboland}, \textit{Nyland}, \textit{Österbotten}, \textit{Birkaland} and \textit{Kymmenedalen}.

\begin{table}[]
\begin{tabular}{|l|c|c|}
\hline
\textit{dialect} & \textit{lines} & \multicolumn{1}{l|}{\textit{words}} \\ \hline
Nyland           & 1903           & 29314                               \\ \hline
Åland            & 1949           & 14546                               \\ \hline
Åboland          & 571            & 9989                                \\ \hline
Österbotten      & 1827           & 30137                               \\ \hline
Birkaland        & 64             & 886                                 \\ \hline
Kymmenedalen     & 64             & 1634                                \\ \hline
\end{tabular}
\caption{Number of lines from each region}
\label{tab:corpus-sentences}
\end{table}

Table \ref{tab:corpus-sentences} shows how many lines of interviews the dataset has from each region. One line represents a turn when a participant speaks and it can consist of a single word or multiple sentences. The word count shows how many words we have for each region. As we can see by comparing the line count and the word count, some regions had participants who spoke considerably more than others.

The data itself is not free of noise, and we take several preprocessing steps. The first step is manually going through the entire dataset and find all cases of characters that are not part of the Swedish alphabets. This revealed that numbers and the word \textit{euro} were consistently normalized by using numbers and the euro sign \textit{€} even though the dialectal transcription had them written out. For example, \textit{nittånhondratrettitvåå} was normalized simply as \textit{1932}. We went through all these cases and wrote the numbers out in standard Swedish, such as \textit{nittonhundratrettitvå}. This is important as we want the model to be able to normalize text, not to convert text into numbers, and any noise in the data would make it harder for the model to learn the correct normalization.

We lowercased the dataset, removed punctuation as they were inconsistently marked between the dialectal transcriptions and their normalizations, and tokenized the data with NLTK \cite{BirdKleinLoper09}. At this point, most of the lines in the corpus had an equal number of words in the transcriptions and their normalizations. However, 1253 lines had a different number of words, as sometimes words are pronounced together but written separately in standard Swedish, such as in the case of \textit{såhäna} that was normalized as \textit{sådana här}. As we want to train our models to operate on token level rather than normalizing full sentences of varying lengths, we need to map the dialectal text to their normalizations on a token level.

For the token level mapping, we train a statistical word aligner model with a tool called Fast align ng \cite{gelling-cohn-2014-simple} which is an improved version of the popular fast align tool \cite{dyer-etal-2013-simple}. Such tools are commonly used in machine translation contexts. We train this aligner with all of our data and use it to align the dialectal and normalized sentences that do not have an equal number of tokens.

We shuffle the data and split it into training and testing. We use 70\% of the data for training the models and 30 \% for evaluation.

\section{Dialect Normalization}

We use a character level NMT model, following the encouraging results achieved with similar architecture over Finnish dialect data \cite{partanen2019dialect}. The advantage of using a character level model is that the model can better learn to generalize the dialectal differences than a word level model, and it can work for out of the vocabulary words as well as it operates on characters instead of words. In practice, when the model is trained, the words are split into characters, and the underscore sign (\textit{\_}) is used to mark the word boundaries. 

We trained different models by varying the length of the input chunk that was given to the model. We trained separate models to predict from one dialectal word at a time to their normalization, two words at a time all the way to five words at a time (see Table \ref{tab:examplestraining} for an example). Providing context is important, as in many situations the correct normalization of a dialectal word cannot be predicted in isolation. At the same time, longer chunks may become harder to learn \cite{partanen2019dialect}, and thereby it is important to find the optimal length that gives the best performance. We use the same random seed\footnote{The seed used is 3435} when training all of the models to make their intercomparison possible.

\begin{table}[]
\begin{tabular}{|l|l|l|}
\hline
           & dialectal text (source)   & normalized text (target)    \\ \hline
chunk of 1 & h u u v u i n t r e s s e & h u v u d i n t r e s s e n \\ \hline
chunk of 3 & k a n \_ j o \_ n å o     & k a n \_ j u \_ n o g       \\ \hline
\end{tabular}
\caption{Examples of training data for different models}
\label{tab:examplestraining}
\end{table}

For all variations in chunk size, we train a character based bi-directional LSTM model \cite{hochreiter1997long} by using OpenNMT-py \cite{opennmt} with the default settings except for the encoder where we use a BRNN (bi-directional recurrent neural network) \cite{schuster1997bidirectional} instead of the default RNN (recurrent neural network) as BRNN has been shown to provide a performance gain in a variety of tasks. 
We use the default of two layers for both the encoder and the decoder and the default attention model, which is the general global attention presented by Luong et al. \cite{luong2015effective}. The models are trained for the default of 100,000 steps.

\section{Results and Evaluation}

We report the results of the different models based on the accuracy and WER (word-error rate) of their prediction when comparing to the gold standard in the test set. WER is a a commonly used metric to evaluate different systems that deal with text normalization and it is derived from Levenshtein edit distance \cite{Levenshtein66Binary} as a better measurement for calculating word-level errors. It takes into account the number of deletions $D$, substitutions $S$, insertions $I$ and the number of correct words $C$, and it is calculated with the following formula:

\begin{equation}
WER = \frac{S + D + I}{S + D + C}
\end{equation}

\begin{table}[]
\begin{tabular}{|l|c|c|}
\hline
                 & \textit{WER} & \textit{accuracy} \\ \hline
no normalization & 76.45        & 23.5\%            \\ \hline
chunk of 1       & 28.58        & 71.4\%            \\ \hline
chunk of 2       & 33.87        & 66.1\%            \\ \hline
chunk of 3       & 93.47        & 14.3\%            \\ \hline
chunk of 4       & 147.24       & 3.6\%             \\ \hline
chunk of 5       & 103.56       & 4.6\%             \\ \hline
\end{tabular}
\caption{Evaluation results of the different models}
\label{tab:evaluationresults}
\end{table}

The results in Table \ref{tab:evaluationresults} show that the best working model is the one that takes only one word into account at a time. This is interesting as earlier research with Finnish shows that the lowest WER is achieved by chunks of 3 words \cite{partanen2019dialect}. The difference in our results is probably due to the fact that we had less training data available for this task, therefore the model worked best in a situation where it did not need to learn a larger context. 
Since the dialect normalization as a task is often heavily dependent of the context, it must be assumed that with enough data the use of larger chunks is beneficial. 

When looking at the results of the best performing model, most of the words look right or have a very minor issue. The most common mistake the model makes is with \textit{ä}, for instance, \textit{teevlingar} gets normalized into \textit{tevlingar}, even though the correct spelling is \textit{tävlingar} (contests). Interestingly, there is some overlap between how \textit{e} and \textit{ä} are pronounced in Swedish, which means that the model would need more data to learn this phenomenon that is not a part of the phonetics of the language, but rather a matter of a spelling convention. Another source of problems are long words, for example, \textit{såmmararbeetare} is normalized into \textit{sommarbetare} instead of \textit{sommararbetare} (summer worker). This type of problems could be solved by introducing a word segmentation model that would split compounds before normalization.

The model trained with chunks of two words at a time has more severe problems with long words as many of them get heaviliy trunkated, for instance, \textit{teevlingar att} becomes \textit{tällev att} instead of \textit{tävlingar att} (contests to), and \textit{i låågstaadie} turns into \textit{i låstade} instead of \textit{i lågstadiet} (in the elementary school). This model makes more mistakes that are more severe with long words than the model trained with one word at a time.

As long words are problematic even for the models of chunks of 1 and 2, it is not surprising that the models trained with longer chunks get even more confused as the length of the input increases. For example \textit{i låågstaadie jåå} is normalized as \textit{iog och då då} by the chunk of 3 model, \textit{och då och} by the chunk of 4 model and \textit{så var var} by the chunk of 5 model. Needless to say, all of these are very wrong.

\section{Conclusions}

Based on previous research, it seemed that having some, but not too much context in normalization was beneficial for the model and improved results. However, in our study, we can conclude that context should be provided for the model only if you can afford it. This means that the more data you have, the longer sequences can be used. But with very little data, it is better to ignore the context and normalize one word at a time, so that the model can learn a better representation of the normative language. As the model can predict top n candidates instead of the top 1 as we did in this research, in the future, it might be interesting to see if contextual disambiguation of normalization candidates can be left to a language model trained only in the normative language.

Our study provides a new important baseline for dialect normalization as a character level machine translation task. 
We show that also a training data that is significantly smaller than previously used can give useful results and decrease the word error rate dramatically. 
It remains as an important question for future research what exactly is the ideal amount and type of training data for dialect normalization. 
Also the variation in linguistic distance between the dialects and orthographies must be one factor that influences the difficulty of the normalization task. 
We have not attempted to evaluate this, but for the further work this could be another useful baseline when we evaluate how well the model can perform under different conditions.

We have published our versions of the training data openly on Zenodo\footnote{https://zenodo.org/record/4060296}, and hope they will play a role in the future endeavors in improving the normalization of Finland Swedish. 
In the future attention should also be paid to normalization challenges of Swedish dialects spoken outside of Finland.


\bibliographystyle{ACM-Reference-Format}
\bibliography{sample-base.bib}


\end{document}